\documentclass[conference]{IEEEtran}
\IEEEoverridecommandlockouts
\usepackage{cite}
\usepackage{amsmath,amssymb,amsfonts}
\usepackage{hyperref}
\usepackage{algorithm}
\usepackage{algpseudocode}
\usepackage{graphicx}
\usepackage{textcomp}
\usepackage{xspace}
\usepackage{array}
\usepackage{ragged2e}
\usepackage{xcolor}
\usepackage{verbatim}
\usepackage{booktabs}

\usepackage{todonotes}

\def\BibTeX{{\rm B\kern-.05em{\sc i\kern-.025em b}\kern-.08em
    T\kern-.1667em\lower.7ex\hbox{E}\kern-.125emX}}
\definecolor{orange}{RGB}{255,119,0}
\definecolor{red}{RGB}{220,0,0}
\definecolor{agreen}{RGB}{74, 198, 148}
\definecolor{purple}{RGB}{158, 62, 177}
\definecolor{darkpurple}{RGB}{170, 70, 210}
\definecolor{aqua}{RGB}{87, 180, 181}
\definecolor{lightblue}{RGB}{72, 123, 232}
\definecolor{hotpink}{RGB}{255, 83, 115}
\definecolor{teal}{RGB}{90, 200, 250}
\definecolor{linkColor}{RGB}{6,125,233}
\definecolor{tomato}{rgb}{1,0.2,0}
\definecolor{grey}{rgb}{0.4,0.4,0.4}

\DeclareRobustCommand{\method}{\mbox{WellFactor}\xspace}

\newcolumntype{C}[1]{>{\centering\arraybackslash}p{#1}}
\newcolumntype{R}[1]{>{\RaggedLeft\arraybackslash}p{#1}}
\newcolumntype{L}[1]{>{\RaggedRight\arraybackslash}p{#1}}
\IEEEpubidadjcol
\IEEEpubid{\begin{minipage}{\textwidth}\ \\[12pt] 979-8-3503-2445-7/23/\$31.00 ©2023 IEEE \end{minipage}}

\begin{document}
\title{\method: Patient Profiling using Integrative Embedding of Healthcare Data}
\author{
\IEEEauthorblockN{
Dongjin Choi\IEEEauthorrefmark{1}, 
Andy Xiang\IEEEauthorrefmark{2}, 
Ozgur Ozturk\IEEEauthorrefmark{2}, 
Deep Shrestha\IEEEauthorrefmark{2}, \\
Barry Drake\IEEEauthorrefmark{3}, 
Hamid Haidarian\IEEEauthorrefmark{2}, 
Faizan Javed\IEEEauthorrefmark{2}, and 
Haesun Park\IEEEauthorrefmark{1}\IEEEauthorrefmark{4}\thanks{\IEEEauthorrefmark{4}Haesun Park is the corresponding author.}
}
\IEEEauthorblockA{\IEEEauthorrefmark{1}\textit{School of Computational Science and Engineering, Georgia Institute of Technology, Atlanta, GA, USA}}
\IEEEauthorblockA{\IEEEauthorrefmark{2}\textit{Kaiser Permanente, USA}}
\IEEEauthorblockA{\IEEEauthorrefmark{3}\textit{Georgia Tech Research Institute, Atlanta, GA, USA}}
\IEEEauthorblockA{\IEEEauthorrefmark{1}\textit{jin.choi@gatech.edu, hpark@cc.gatech.edu}}
\IEEEauthorblockA{\IEEEauthorrefmark{2}\textit{\{Andy.X.Xiang, Ozgur.X.Ozturk, deep.x.shrestha, Hamid.Haidarian, Faizan.X.Javed\}@kp.org}}
\IEEEauthorblockA{\IEEEauthorrefmark{3}\textit{drakeleeb@gmail.com}}
}


\maketitle

\begin{abstract}
In the rapidly evolving healthcare industry, platforms now have access to not only traditional medical records, but also diverse data sets encompassing various patient interactions, such as those from healthcare web portals.
To address this rich diversity of data, we introduce \method: a method that derives patient profiles by integrating information from these sources.
Central to our approach is the utilization of constrained low-rank approximation. \method is optimized to handle the sparsity that is often inherent in healthcare data.
Moreover, by incorporating task-specific label information, our method refines the embedding results, offering a more informed perspective on patients.
One important feature of \method is its ability to compute embeddings for new, previously unobserved patient data instantaneously, eliminating the need to revisit the entire data set or recomputing the embedding. Comprehensive evaluations on real-world healthcare data demonstrate \method's effectiveness.
It produces better results compared to other existing methods in classification performance, yields meaningful clustering of patients, and delivers consistent results in patient similarity searches and predictions.
\end{abstract}

\section{Introduction}
\label{sec:intro}
The digital revolution and the rise of healthcare web portals have significantly changed patient interactions within the healthcare domain.
While these portals primarily serve purposes such as accessing personal clinical records or scheduling appointments, they also store extensive data on patient activities, including searches for clinical information and browsing patterns~\cite{tanbeer2021myhealthportal,coughlin2017patient,chou2002healthcare,moody2005health}.
This surge in healthcare big data comes with challenges. One such challenge is the infrequent nature of user interactions on these portals.
While these portals experience infrequent user interactions, unlike generic web domains, they maintain a more accurate record of user demographics and offer detailed clinical diagnosis information.

Profile-based models, leveraging user interaction histories, have been effectively used for recommendation across various domains~\cite{DBLP:conf/ictai/PretschnerG99,DBLP:conf/cikm/ShenTZ05,DBLP:conf/www/SugiyamaHY04,DBLP:conf/kdd/TanSZ06}.
Studies on embedding often focus on using the learned vector in downstream machine learning models regardless of their numerical value~\cite{DBLP:conf/nips/MikolovSCCD13}. In contrast, our approach, akin to profile-based models, recognizes the latent significance of each dimension in the learned profiles for further recommendations.
This is demonstrated in our \textit{Cluster Analysis} (\autoref{subsec:cluster-analysis}), where embedding dimensions directly correspond to key patient cluster characteristics.
\begin{figure}[tbp]
  \centering
  \includegraphics[width=0.47\textwidth]{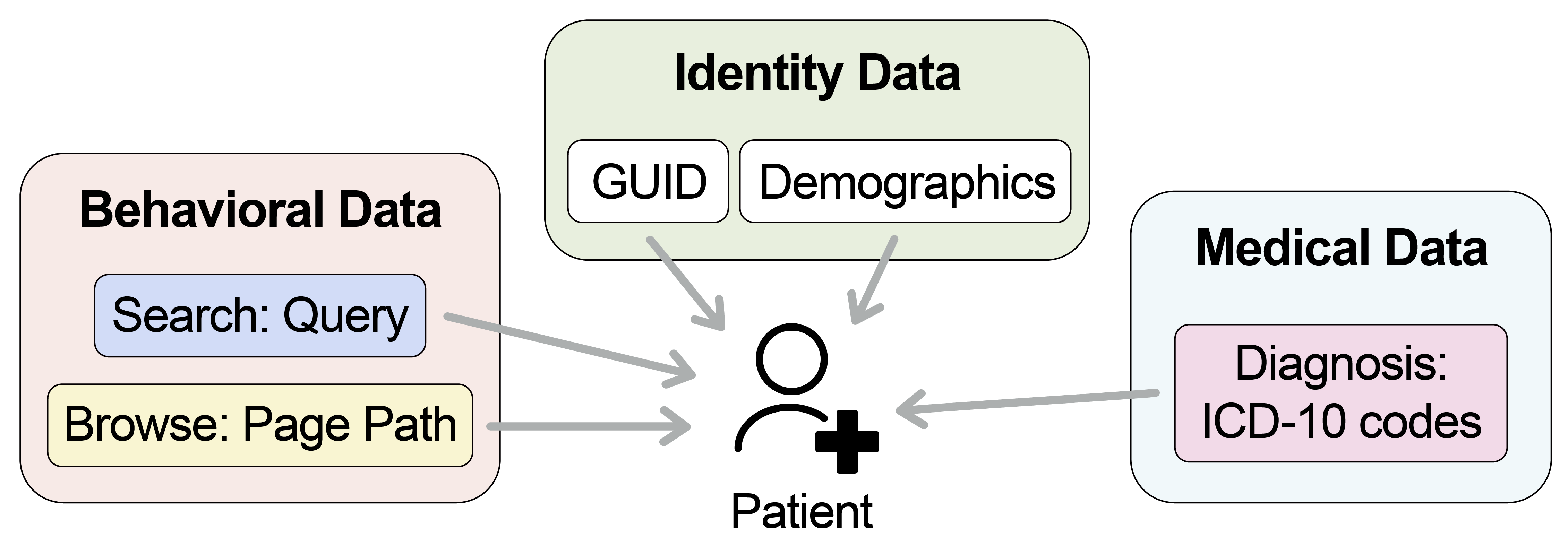}
  \caption{Illustration of the diverse patient data sources collected from interactions on the web portal and with medical professionals.}
  \label{fig:data_formulation}
  \vspace{-6mm}
\end{figure}

This study aims to create comprehensive patient profiles on healthcare web portals and utilizes demographic data for refined classification and recommendation tasks. These profiles integrate interaction data with diagnostic information, offering a robust foundation for various applications, including personalized healthcare recommendations. 
While many digital platforms aim solely to enhance user engagement, healthcare portals have additional considerations.
Consequently, every patient embedding or subsequent recommendation must reflect both patient preferences and prospective health advantages.
For instance, demographic nuances, such as age, can significantly influence user interaction patterns. Younger users, notably those in their 20s and 30s, may display a pronounced affinity for mobile applications, making them more receptive to app-based health recommendations~\cite{hwang2016factors,kurniawan2008older,mitzner2010older,hall2015digital}.
Conversely, older users, particularly those in their 50s and above, may also require more mental wellness support due to chronic illnesses and related pains~\cite{fiske2009depression,chhabra2018smartphone,irvine2015mobile}.

In this study, we introduce \method, an algorithm based on constrained low rank approximation (CLRA) that particularly employs nonnegativity constraints. This approach facilitates the integration of diverse user feature vectors directly within the objective function.
Several innovative aspects define \method:
\begin{itemize}
    \item \textbf{Integration of data}: The incorporation of integrative objective function and nonnegativity constraints, \method seamlessly blends various data sources, creating a comprehensive representation of patient profiles. Moreover, our method handles missing or unobserved data domains, addressing few-shot scenarios.
    \item \textbf{Efficient embedding computation}: \method utilizes an alternating block coordinate descent algorithm optimized for unique characteristics of the objective function. It avoids materializing extensive data matrices. Moreover, \method predicts embeddings for previously unseen patients without the need to re-examine the entire data.
    \item \textbf{Task-specific embeddings}: With the incorporation of semi-supervision, \method is tailored to produce high-quality embeddings, particularly optimized for certain tasks. The enhanced embeddings offer a more refined view of patients.
\end{itemize}

We evaluated \method using real-world datasets obtained from Kaiser Permanente's web portal. Our evaluation process has shown the effectiveness of \method compared to other methods. The results consistently showed that \method outperformed other existing methods, particularly in terms of classification accuracy. 
The method demonstrated its capacity to generate meaningful patient clusters with the potential to tailor personalized healthcare recommendations.
Moreover, its efficacy in patient similarity searches and predictive tasks showcases its comprehensive capability to utilize and represent patient data.
Our results indicate the effectiveness of \method in handling healthcare data and its potential use for healthcare professionals and researchers.
\section{Related Work}
As described in Section \ref{sec:intro}, the recommendation of wellness apps in healthcare web portals is related to several areas such as content-based recommendation, prediction in the healthcare domain, clustering, and nonnegative matrix factorization. We offer a brief literature review on each of these topics related to our proposed approach.

\subsection{Content-based Recommendation}
Our goal is to produce personalized recommendations tailored to individual user preferences and needs. Such methodologies have been explored under content-based recommendation. Techniques that rely on the representation of item contents that align with user interests are the basis of this approach. Profile-based models emphasize the formation of user preferences and interests as vectors or lists. Some models construct user profiles based on users' search or browsing history within a specific time window \cite{DBLP:conf/www/SugiyamaHY04, DBLP:conf/kdd/TanSZ06}.
A clustering-based approach groups user trajectories into distinct clusters, suggesting different recommendations for each group~\cite{linden2003amazon}. Another approach treats recommendations as a classification problem \cite{DBLP:conf/aaai/BasuHC98}.

Several popular techniques are based on Artificial Neural Networks (ANNs), with some incorporating aspects of the Markov Decision Process \cite{sutton1998introduction}. For instance, some methods employ reinforcement learning techniques to track the evolving interests of users \cite{liao2022deep, DBLP:conf/aaai/ZhaoGZYLTL21}. Others, such as the self-attention-based method \cite{chen2022time}, utilize a time embedding model.

\subsection{Contextual Embedding Methods and Clinical Domain-Specific Embeddings}
Contextual embedding methods generate vector representations of text that capture semantic nuances. GPT-2 (Generative Pre-trained Transformer 2) \cite{radford2019language} is a method used for capturing long-term textual dependencies. SentenceBERT (SBERT) \cite{reimers2019sentencebert} produces a fixed-size embedding vector for an entire text input.
Beyond generic embeddings, the healthcare domain demands specialized embedding techniques to understand the semantics of medical text. BioSentVec is one such approach which focuses on matrix factorization-based embeddings for biomedical sentences~\cite{chen2019biosentvec}.

\subsection{Prediction in the Healthcare Domain}
The method we introduce is related to recommendations within healthcare web portals. Previous research, such as KETCH \cite{cui2022ketch}, has explored recommending relevant threads on healthcare forums based on user symptoms or conditions. The active incorporation of user diagnosis data in our study is inspired by such insights. Our research direction is motivated by representation techniques, such as Metacare++ \cite{tan2022metacare++}, that leverage the hierarchical structure of ICD-9 diagnosis codes.

\subsection{Clustering and Constrained Low Rank Approximation}
Our profiling technique is inspired by clustering methods, especially those focused on constrained low rank approximation \cite{du2019hybrid, kim2019topicsifter}. The proposed method yields low-dimensional representations, interpretable as probability distributions. Clustering methods have been applied in recommendation systems, including session clustering for web page recommendations \cite{demir2010multiobjective} and user clustering based on time-framed data \cite{wang2004effective}.
Clustering based on evolving browsing data has also been used for recommendations \cite{li2021novel}. Recommendations have been improved using multi-view clustering techniques \cite{guo2015leveraging}. Matrix factorization-based techniques have been employed to produce transparent recommendations \cite{melchiorre2022protomf}. The MEGA model \cite{whang2020mega} uses nonnegative matrix factorization with a wide range of hypergraphs.
\section{Healthcare Web Portal Data}
\label{sec:data}
In this study, our primary focus is on the integration of clinical data with internet activity data within the healthcare domain. Given the proprietary nature of our primary dataset, we detail its characteristics to enhance reproducibility and discuss its potential applicability to analogous public datasets.
The data we used in our research, sourced from Kaiser Permanente, has been anonymized in accordance with the Health Insurance Portability and Accountability Act (HIPAA)\footnote{Health Insurance Portability and Accountability Act (\url{https://www.hhs.gov/hipaa/})}, ensuring the privacy and security of patients' information.
\begin{figure*}[tb]
  \centering
  \includegraphics[width=0.97\textwidth]{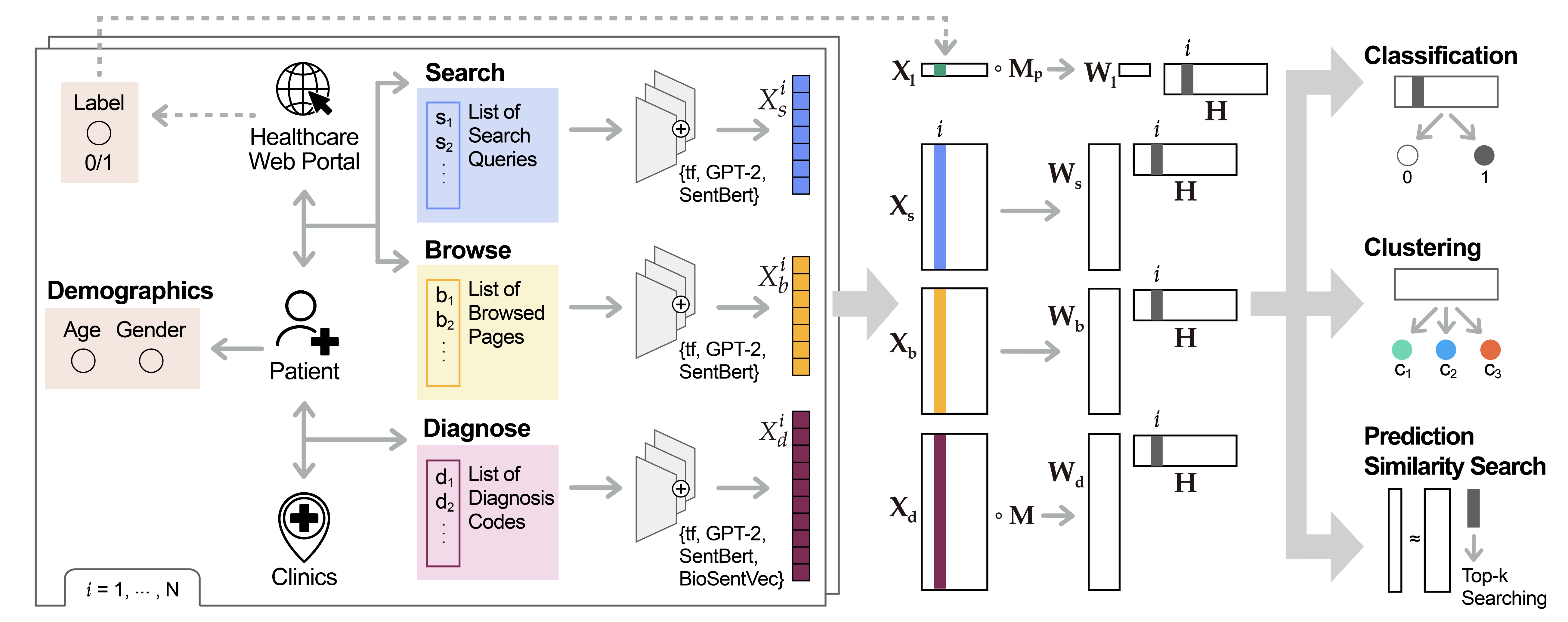}
  \caption{Graphical overview of the proposed \method patient profiling framework.}
  \label{fig:overview}
\end{figure*}

\begin{table}[t]
\centering
\caption{Statistics on the Data Set Utilized in Our Study}
\label{tab:data-stats}
\setlength{\tabcolsep}{0.4em} 
\begin{tabular}{L{1.30cm}R{1.28cm}R{1.28cm}R{1.6cm}R{1.7cm}}
\toprule
 & \textbf{Diagnosis} & \textbf{Search Histories} & \fontsize{8pt}{0pt}\selectfont\textbf{Browsing Activities} & \fontsize{8pt}{0pt}\selectfont\textbf{Demographics}\\
\midrule
\fontsize{8pt}{0pt}\selectfont\# Patients & $599,499$ & $298,574$ & $636,150$ & $1,177,031$ \\
\fontsize{8pt}{0pt}\selectfont\# Instances & $6,268,788$ & $1,220,210$ & $299,099,939$ & $1,177,031$ \\
\fontsize{8pt}{0pt}\selectfont\# Fields & $3$ & $1$ & $1$ & $2$ \\
\bottomrule
\end{tabular}
\end{table}

The data capture patient interactions on the Kaiser Permanente Digital (KPD) website\footnote{\url{https://healthy.kaiserpermanente.org}}. As depicted in \autoref{fig:data_formulation}, this dataset offers a detailed view of patient information. It encompasses medical diagnoses based on ICD-10 codes~\cite{world1992icd} gathered during interactions with medical professionals, and a broad spectrum of digital user interactions: search histories, browsing activities, and demographic particulars such as age and gender.
All information, especially medical diagnoses, is anonymized to ensure confidentiality. For a more granular overview of the dataset, refer to \autoref{tab:data-stats}. 

\begin{table}[t]
\centering
\caption{Overview of Data Fields and Examples}
\label{tab:data-example}
\begin{tabular}{L{1.4cm}C{1.8cm}R{4.0cm}}
\toprule
\textbf{Data Used} & \textbf{Field Name} & \textbf{Examples} \\
\midrule
Diagnosis & ICD-10 & R80.9 \\
 & official text & \fontsize{7pt}{0pt}\selectfont\textit{PROTEINURIA} \\
 & friendly text & \fontsize{7pt}{0pt}\selectfont\textit{PROTEINURIA (PROTEIN IN URINE)} \\
\midrule
Search Histories & query & \textit{How to eat low carb}\\
\midrule
Browsing Activities & site-path & \textit{`kporg:health-wellness:healtharticle.40-positive-affirmations'} \\
\midrule
Demographics & age & 48 \\
 & gender & male \\
\bottomrule
\end{tabular}
\end{table}

\autoref{tab:data-example} provides an overview of the data sets, their fields, and examples for each data set. 
Within the diagnosis data, the primary field is the ICD-10 codes. These codes are derived from treatments patients receive at Kaiser Permanente (KP) and its affiliated clinics. Confirmed diagnoses from these interactions are archived in KP's electronic health record systems. Each diagnosis in KP's electronic health record systems includes an ``official text" (medical terminology) and a ``friendly text" (more colloquial name) annotated by KP specialists for internal reference.

The data set also contains patients' search histories on the KP web portal. This platform allows users to input queries to locate appointments, articles, and medical providers. These search results are presented similarly to a conventional search engine results page (SERP), and we specifically collected the query expressions patients entered.
Moreover, the browsing activity data set contains the patients' web page interaction information on both desktop and mobile platforms. This extends beyond search results to include pages displaying personal data, clinician specifics, and appointment-booking locations within the KP framework. 
We gathered information on web pages accessed by users, which are marked with the path signifying its position in KP's internal web hierarchy. Furthermore, essential demographics, such as age and gender, were also recorded in the system.
For all data, we limited the data period of 2022 (from 2022-01-01 to 2022-12-31)

\subsection{Label Collection for Semi-supervision and Evaluation}
\label{subsec:label_collection}
For semi-supervision and evaluation, we sourced labels based on users' interactions with a mental health app banner displayed on KP's home page during 2022.
These apps include Calm, Ginger, and myStrength.
A cohort of randomly selected users during this timeframe was exposed to this banner, which directed them to download the self-care apps. Our labels specifically identify whether a user clicked on this banner.
After label collection, we found that a mere 15,382 users interacted with the banner, a small portion when compared with the total number of patients (1.1 M). Given this significant imbalance, undersampling will be adopted in subsequent experiments, such as classification.

\subsection{Relation to Public Datasets}
\label{subsec:relation_public_datasets}
Our research dataset integrates clinical and digital domain records for over a million individuals, differing from datasets like MIMIC-IV~\cite{johnson2020mimic}, which primarily offers patient anonymized clinical notes and demographics. The integrative nature of our dataset underscores its value and the opportunities it presents, even when public data is favored for reproducibility. As public datasets evolve, aligning them with diverse data sources will make them more valuable. A future direction involves integrating textual data with public clinical datasets like MIMIC-IV.
\section{Methodology}
Our proposed patient profiling method integrates multiple sources of data, distinguishing it from existing methods. By integrating information from patients' digital interactions and medical records using our algorithms, our approach aims to provide a broader understanding of patients. 
This is further enhanced by the employment of semi-supervision techniques.
The architecture of our method is presented in \autoref{fig:overview}.

Leveraging the comprehensive dataset detailed in \autoref{sec:data}, our methodology makes full use of all user data present in the digital healthcare platform. Digital healthcare platforms typically provide user interaction data with digital interfaces, akin to the e-commerce sector, and diagnostic data from user interactions with medical experts. We propose algorithms that can effectively integrate multiple sources of heterogeneous data domains, producing more accurate patient profiles. 
While the details of our algorithms and implementations are demonstrated using the search, browsing, and diagnosis data views from \autoref{sec:data}, our method can be adapted to settings with any number of heterogeneous data domains.

In this section, we introduce a new approach for learning user profiles that utilizes multiple data sources simultaneously.
The features obtained from the proposed method provide a richer representation of user behavior and preferences, as they integrate information from various sources, such as content, semantic relationships, and domain-specific information.
Some of the commonly utilized methods for information fusion include early fusion which merges raw data at the
data representation level~\cite{lahat2015multimodal} and late fusion which solves a given problem applying solution methods 
separately to each data set and then merges the results~\cite{atrey2010multimodal}.
Our method integrates the objective functions from all data sets into a single objective function.

The objective function level information fusion method we introduce here does not require an 
input matrix that
contains all merged raw data as in early fusion. Instead, each part of the merged objective function takes
each representation of a data set as its input.
Then the goal of the overall merged objective function is to find one common lower-dimensional embedding that captures the essential information from all {\em views} of the data by computing a common low rank factor.

\subsection{Feature Processing}
\label{subsec:featureprocessing}
In order to provide effective recommendations, we consider various user features derived from collected data.
We utilize the content information representing the content in the standard Term Frequency (TF) encoding.
In addition, we incorporate features that capture semantic relationships between words and phrases within the user's textual data. We use GPT-2~\cite{radford2019language}, a generative language model, and sentenceBERT~\cite{reimers2019sentencebert}, a variation of the BERT model, optimized for sentence-level representations, to process all data types, including search, browsing, and diagnosis records. 
For the diagnosis data, we additionally utilize BioSentVec~\cite{chen2019biosentvec}, a sentence embedding model specifically trained on biomedical texts, to capture the domain-specific semantic information more effectively.
Since the range of elements in these three matrices from different data sets may vary significantly,
we use Min-Max scaling~\cite{DBLP:journals/pr/JainNR05}
for each matrix.

\subsection{Learning User Profiles}
To illustrate the details, we assume that we have the three different views for a set of $n$ users, i.e., the users' web search data, browsing data, and  diagnosis data. Then we generate three feature-by-user matrices, which are
search-by-user matrix $\mathbf{Y}_s \in {\bf R}^{m_s \times n}$, 
browsing-by-user matrix $\mathbf{Y}_b  \in {\bf R}^{m_b \times n}$, 
and diagnosis-by-user matrix  $\mathbf{Y}_d \in {\bf R}^{m_d \times n}$. 
Given a matrix 
$\mathbf{Y}_i$ where $i \in \{s, b, d\}$, the scaled matrix $\mathbf{X}_i$ is obtained as 
$$ \mathbf{X}_i = ({\mathbf{Y}_i - \min(\mathbf{Y}_i)})/({\max(\mathbf{Y}_i)-\min(\mathbf{Y}_i)}), $$    
where $\min(\mathbf{Y}_i)$ denotes the matrix of the same size as ${\mathbf{Y}_i}$ where all elements are identically set to the minimum of $(\mathbf{Y}_i)$,
and $\max(\mathbf{Y}_i)$ is defined analogously using
the maximum value of the elements in $\mathbf{Y}_i$.

If we were to find a low rank approximation for just one of the data sets $\mathbf{X}_i$ via 
nonnegative matrix factorization (NMF) \cite{kim2014algorithms}, 
then the objective function would be 
$$\min_{\{\mathbf{W_i}, \mathbf{H}\} \geq 0} \left\|\mathbf{Y}_i-\mathbf{W}_i\mathbf{H}\right\|_F. $$
Now we merge three objective functions and produce a common embedding.
Then the objective function for this method is as follows:
\begin{align}
\min _{(\mathbf{W}_s, \mathbf{W}_b, \mathbf{W}_d, \mathbf{H})\geq 0} 
&\alpha_s\left\|\mathbf{X}_s-\mathbf{W}_s \mathbf{H}\right\|_F^2 +
\alpha_b\left\|\mathbf{X}_b -\mathbf{W}_b\mathbf{H}\right\|_F^2 \nonumber\\
&+\alpha_d\left\|\mathbf{X}_d-\mathbf{W}_d \mathbf{H}\right\|_F^2,
\label{eqn:merged}
\end{align}
where $\alpha_s$, $\alpha_b$, $\alpha_d$ denote balancing factors for each low-rank approximation term, and $\mathbf{X}_s$, $\mathbf{X}_b$, $\mathbf{X}_d$ represent the feature by data matrices from each domain: search, browsing, and diagnosis, respectively. Note that the factor $H$ is common across all domains, which provides an embedding in $k$ dimensional space for the data items that reflects their relationships with search, browsing, and diagnosis, simultaneously.
The factors $\mathbf{W}_s$, $\mathbf{W}_b$, and $\mathbf{W}_d$ represent the basis matrices in the reduced $k$ dimensional space for each domain.

The matrix $\mathbf{H}$, resulting from solving our merged objective function,
serves a dual purpose: soft clustering and embedding.
Each column within $\mathbf{H}$, specifically the $i$\textsuperscript{th} column $H^i$, encapsulates an integrated embedding of the $i$\textsuperscript{\textit{th}} data item.
Furthermore, it can also be understood as a probability distribution that illustrates how the $i\textsuperscript{th}$ data item spans across clusters.
This approach facilitates a clearer interpretation of results, as the learned embedding can be understood in terms of the contributions of different cluster representatives present in the columns of the basis matrices $\mathbf{W}_s$, $\mathbf{W}_b$, and $\mathbf{W}_d$.
This methodology not only uses user profiles for downstream tasks but also provides a latent factor representation, unifying characteristics across all domains.
For a more in-depth exploration of the interpretation of the factor matrix $\mathbf{H}$ within the context of the soft clustering, we refer to \cite{kim2014algorithms}.

To optimize the objective function in~\autoref{eqn:merged}, we adopt a Block Coordinate Descent (BCD) approach.  
In each iteration of our proposed BCD method, we alternate updating one of the matrices $\mathbf{W}_s$, $\mathbf{W}_b$, $\mathbf{W}_d$, and $\mathbf{H}$ while fixing the other three matrices by solving the following subproblems until a stopping criteria is satisfied:
\begin{equation}
\begin{split}
& \min_{\mathbf{W}_i \geq 0} \left\|\mathbf{X}_i - \mathbf{W}_i\mathbf{H} \right\|_F, \mbox{for} \quad i = s, b, d\\ 
&\min_{\mathbf{H} \geq 0} \left\|
\begin{bmatrix} \sqrt{\alpha_s}\mathbf{X}_s \\
\sqrt{\alpha_b}\mathbf{X}_b \\
\sqrt{\alpha_d}\mathbf{X}_d \end{bmatrix} - \begin{bmatrix} \sqrt{\alpha_s}\mathbf{W}_s \\
\sqrt{\alpha_b}\mathbf{W}_b \\
\sqrt{\alpha_d}\mathbf{W}_d \end{bmatrix}\mathbf{H}\right\|_F.
\label{eqn:optimize_H}
\end{split}
\end{equation}
Each of the four subproblems is a nonnegativity-constrained least squares (NLS) problem and there are several methods that can effectively solve these NLS problems. We utilize the BPP (Block Principal Pivoting) method as it has been shown to produce the best performance in previous extensive studies and for various possible stopping criteria, see \cite{kim2014algorithms}. Assuming that each subproblem has a unique solution, the limit point of the iteration will be a stationary point ~\cite{kim2014algorithms,bertsekas1997nonlinear}.

In fact, the same objective function can be expressed as
\begin{equation}
\min_{(\tilde{\mathbf{W}}, \mathbf{H}) \geq 0}
\left\|
\begin{bmatrix} 
\sqrt{\alpha_s}\mathbf{X}_s \\
\sqrt{\alpha_b}\mathbf{X}_b \\
\sqrt{\alpha_d}\mathbf{X}_d 
\end{bmatrix} 
- \tilde{\mathbf{W}} \mathbf{H} \right\|_F,
\label{eqn:earlyfusion}
\end{equation}
which is the same as applying an NMF to the 
$\mathbf{X}_i$'s scaled by $\sqrt{\alpha_i}$ and stacking them up. Then by scaling and partitioning the computed factor 
$\mathbf{\tilde{W}}$, we obtain the  $\mathbf{W}_i$'s.
However, there are added advantages of the proposed objective function level fusion, in terms of its 
generalizability.
The first is when some view of the data is in the form of data-data relationships. For example, suppose we have an additional view of the data representing the relationships or interactions among the users, which is represented in a similarity matrix $\mathbf{A}$. Then we can add 
an additional term of Symmetric NMF \cite{kuang2012symmetric,kuang2015symnmf}, 
$\alpha_a\left\|\mathbf{A}-\mathbf{H}^T \mathbf{H}\right\|_F$
to \autoref{eqn:merged} and compute the common $\mathbf{H}$ factor that represents all four views of the data. This data-data relationship information cannot be represented in the form in \autoref{eqn:earlyfusion}.
In addition, when there are some missing elements in a data matrix,
we can compute the common factor $\mathbf{H}$ bypassing the missing elements, i.e.,
not letting the missing elements influence the result.  
We discuss this in detail in the next section for \autoref{eqn:merged}. 

\subsection{Unobserved Data: Assumptions and Matrix Masking}
Data and features in each domain are not always fully observed. 
It is essential to develop a method to handle unobserved or missing data and features. This approach aids in producing accurate recommendations based on the available information.
For search and browsing data, we assume a closed-world assumption~\cite{DBLP:conf/adbt/Reiter77}, meaning that unobserved matrix entries indicate no existing relationship.
This is due to users having the freedom to search and browse, and they can also choose not to engage in such activities based on their intentions.
On the other hand, for diagnosis data, we utilize an open-world assumption~\cite{DBLP:journals/pieee/Nickel0TG16}, i.e., unobserved matrix entries are considered to represent an {\em unknown} relationship.
This is because the absence of a diagnosis does not always reflect a user's intention. Instead, it may indicate that the user has not received a diagnosis from a medical expert.

In order to handle unobserved entries in the diagnosis data, we introduce a masking matrix $\mathbf{M}\in \{0,1\}^{m\times n}$, where its entry is 1 when the corresponding entry in the diagnosis matrix $\mathbf{X}_d$ is observed and 0 when it is not observed.
We modify the objective function in~\autoref{eqn:merged} to incorporate the masking matrix as follows 
\begin{align}
\min _{(\mathbf{W}_s, \mathbf{W}_b, \mathbf{W}_d, \mathbf{H})\geq 0} &\alpha_s\left\|\mathbf{X}_s-\mathbf{W}_s \mathbf{H}\right\|_F^2 +
\alpha_b\left\|\mathbf{X}_b-\mathbf{W}_b\mathbf{H}\right\|_F^2 \nonumber\\
&+
\alpha_d\left\|\mathbf{M} \circ (\mathbf{X}_d-\mathbf{W}_d \mathbf{H})\right\|_F^2.
\label{eqn:merged_mask}
\end{align}
As in the previous section, we use the BCD framework to solve \autoref{eqn:merged_mask}, updating the four factor matrices in each iteration. Updating of $\mathbf{W}_s$ and $\mathbf{W}_b$ can be done in the same way as in \autoref{eqn:optimize_H}, respectively. However, the updating of $\mathbf{H}$ and $\mathbf{W}_d$ will be different due to the masking matrix.
Considering the effects of the masking matrix, the subproblems in \autoref{eqn:optimize_H} change as follows:
\begin{align*}
&\min_{\mathbf{W}_d \geq 0} \left\|
\mathbf{M} \circ
\left(
\mathbf{X}_d -  \mathbf{W}_d \mathbf{H}
\right)\right\|_F, \\ 
& \min_{\mathbf{H} \geq 0} \left\|
\begin{bmatrix} \sqrt{\alpha_s}\mathbf{X}_s \\
\sqrt{\alpha_b}\mathbf{X}_b \\
\mathbf{M} \circ 
\left(\sqrt{\alpha_d} \mathbf{X}_d \right) \end{bmatrix} - \begin{bmatrix} \sqrt{\alpha_s}\mathbf{W}_s \mathbf{H}\\
\sqrt{\alpha_b}\mathbf{W}_b \mathbf{H}\\
\mathbf{M} \circ \left( \sqrt{\alpha_d}\mathbf{W}_d \mathbf{H} \right)\end{bmatrix}\right\|_F.
\end{align*}
Accordingly, $\mathbf{W}_d$ and $\mathbf{H}$ can be updated row by row and column by column, respectively, using the following rules:
\begin{align}
&
\min_{\mathbf{W}_d(j, :) \geq 0}\left\|
\mathbf{X}_d(j, :)D(\mathbf{M}(j, :))
-
\mathbf{W}_d(j, :)\mathbf{H} D(\mathbf{M}(j, :))\right\|_F , \label{eqn:wd_update} \\
&\min_{\mathbf{H}(:, j) \geq 0}
\scalebox{0.85}{$
\left\|
\begin{bmatrix} \sqrt{\alpha_s}\mathbf{X}_s(:, j) \\
\sqrt{\alpha_b}\mathbf{X}_b(:, j) \\
\sqrt{\alpha_d}
D(\mathbf{M}(:, j)) 
\mathbf{X}_d(:, j) \end{bmatrix} - \begin{bmatrix} \sqrt{\alpha_s}\mathbf{W}_s\\
\sqrt{\alpha_b}\mathbf{W}_b\\
\sqrt{\alpha_d}
D(\mathbf{M}(:, j)) \mathbf{W}_d\end{bmatrix}\mathbf{H}(:, j)\right\|_F$}, \label{eqn:h_update}
\end{align}
where $D(\mathbf{z})$ denotes the diagonal matrix whose diagonal entries are defined by the given row or column vector $\mathbf{z}$, i.e., $D(\mathbf{z})=\left[d_{i j}\right]_{n \times n}$ and $d_{i i}=z_i$ where $\mathbf{z} \in \mathbb{R}^n$ and $z_i$ is the $i$-th component of $z$.

\subsection{Semi-Supervised Embedding}

Incorporating partially known prior information can enhance the quality of patient profiles. Our semi-supervised learning approach, grounded in data-level supervision, exploits the partially observed labels of the data items to refine the embedding quality.
This semi-supervised approach optimizes the patient profiling framework's efficacy, potentially leading to improved outcomes in applications.

Given the availability of partial label information for data items, this prior knowledge can be seamlessly integrated into the embedding algorithm using the label matrix $\mathbf{X}_l$.
The matrix $\mathbf{X}_l \in \{0,1\}^{p \times n}$ denotes the partially observed labels where $p$ represents the number of distinct label classes for users.
An entry $x_{l}^{ij} = 1$ indicates that the $j$\textsuperscript{th} object is part of the $i$\textsuperscript{th} class.

In our application context, a data item represents a user. As highlighted in Section \ref{sec:data}, we have label information indicating whether a user has accessed any mental wellness support apps, particularly those focused on self-care such as Calm, Ginger, and myStrength. It is worth noting that while we exemplify the semi-supervision process using labels of mental wellness support app download attempts in this context, the label could be representative of various patient statuses. This might include particular diagnoses or interactions with specific medical resources, emphasizing the flexibility of our approach.

By integrating this partially observed label information into our semi-supervised learning framework, the algorithm makes use of the available information to capture the underlying structure of the user. Consequently, this contributes to more informed patient profiles.
To implement this semi-supervision, we extend the objective function by including a new term that minimizes the Frobenius norm of the difference between the observed labels $\mathbf{X}_l$ and their approximated values $\mathbf{W}_l\mathbf{H}$, where $\mathbf{W}_l$ is an additional basis matrix for labels:
\begin{equation*}
\begin{split}
&\min _{(\mathbf{W}_s, \mathbf{W}_b, \mathbf{W}_d, \mathbf{W}_l, \mathbf{H})\geq 0} \alpha_s\left\|\mathbf{X}_s-\mathbf{W}_s \mathbf{H}\right\|_F^2 +
\alpha_b\left\|\mathbf{X}_b-\mathbf{W}_b\mathbf{H}\right\|_F^2 \\
&+ \alpha_d\left\|\mathbf{M} \circ (\mathbf{X}_d-\mathbf{W}_d \mathbf{H})\right\|_F^2 +
\alpha_l\left\|\mathbf{M}_l \circ (\mathbf{X}_l-\mathbf{W}_l \mathbf{H})\right\|_F^2,
\end{split}
\end{equation*}
where $\alpha_l$ is a regularization parameter that controls the trade-off between fitting the observed labels and the other components of the objective function, and $\mathbf{M}_l$ is an entry-wise masking matrix that indicates whether an entry in $\mathbf{X}_l$ is observed or not.

The update process for $\mathbf{W}_l$ is analogous to the columnwise update of $\mathbf{W}_d$ as shown in~\autoref{eqn:wd_update}.
Therefore, the detailed procedure will be skipped for brevity. The columnwise update of $\mathbf{H}$ can be extended from~\autoref{eqn:h_update} and is expressed as follows:
\begin{equation*}
\min_{\mathbf{H}(:, j) \geq 0}
\scalebox{0.8}{$
\left\|
\begin{bmatrix} \sqrt{\alpha_s}\mathbf{X}_s(:, j) \\
\sqrt{\alpha_b}\mathbf{X}_b(:, j) \\
\sqrt{\alpha_d}
D(\mathbf{M}(:, j)) 
\mathbf{X}_d(:, j) \\
\sqrt{\alpha_p}
D(\mathbf{M}_p(:, j)) 
\mathbf{P}(:, j) \end{bmatrix} - \begin{bmatrix} \sqrt{\alpha_s}\mathbf{W}_s\\
\sqrt{\alpha_b}\mathbf{W}_b\\
\sqrt{\alpha_d}
D(\mathbf{M}(:, j)) \mathbf{W}_d\\
\sqrt{\alpha_P}
D(\mathbf{M}_p(:, j)) \mathbf{W}_p\end{bmatrix}\mathbf{H}(:, j)\right\|_F$}
\end{equation*}

\subsection{Embedding of Previously Unseen Data Items}
Given the multi-domain information about users, 
we can predict the embedding for a previously unseen patient based on the already computed bases vectors for search, browsing, and diagnosis, i.e., $\mathbf{W}_s$, $\mathbf{W}_b$, and $\mathbf{W}_d$. In the following, we show how to achieve this using the integrated nonnegative least squares (NNLS) method.
Suppose a new patient signs up and we have information on the person such as search histories, browsing activities, and diagnosis.
Using the results we have already computed from the \method method from known patients, we can determine a profile for this new patient.
Let $q$ be the patient's ID to be entered as a query.
The patient's observed search, browse, and diagnosis records, once processed analogously as described in~\autoref{subsec:featureprocessing}, are represented as column vectors $X_s^q$, $X_b^q$, and $X_d^q$.
Consequently, computing for $\mathbf{H}^q$ in the equation below offers the patient's profile representation across three subspaces represented in $\mathbf{W}_s$, $\mathbf{W}_b$, and $\mathbf{W}_d$:
\begin{equation}
\begin{split}
\min _{H^q\geq 0} 
&\alpha_s\left\|X_s^q-\mathbf{W}_s H^q\right\|_F^2 +
\alpha_b\left\|X_b^q -\mathbf{W}_b H^q\right\|_F^2 \\&+\alpha_d\left\|X_d^q-\mathbf{W}_d H^q\right\|_F^2.
\label{eqn:prediction}
\end{split}
\end{equation}
The objective function in \autoref{eqn:prediction} computes the patient embedding, $\mathbf{H}^q$. This embedding succinctly represents the patient based on their individual records, placing them within the established embedding subspace.

To foster reproducibility and encourage further developments in patient profiling, the code for our \method framework has been made publicly available.
Interested researchers and developers can access and utilize the codebase via our GitHub repository\footnote{\url{https://github.com/skywalker5/wellfactor}}.
\begin{table*}[th]
\centering
\caption{
Comparison of various embedding methods evaluated using XGBoost: Metrics include ROC-AUC, Accuracy, Recall, Precision, and F1-score. Results are presented in terms of mean percentages with standard deviations. The \textbf{bold} values represent the best scores and \underline{underlined} values signify the second-best scores for each metric.
}
\label{tab:auc-scores-xgboost}
\begin{tabular}{lrrrrr}
\toprule
\textbf{Method} & \textbf{ROC-AUC} & \textbf{Accuracy} & \textbf{Recall} & \textbf{Precision} & \textbf{F1-score} \\
\midrule
HashGNN & 65.95 $\pm$ 0.63 & 62.09 $\pm$ 0.44 & 65.91 $\pm$ 3.84 & 61.60 $\pm$ 1.20 & 63.61 $\pm$ 1.56 \\
GPT-2 (PCA) & 76.73 $\pm$ 0.52 & 69.74 $\pm$ 0.58 & 69.92 $\pm$ 3.53 & 70.11 $\pm$ 1.79 & 69.94 $\pm$ 1.26 \\
SentBERT (PCA) & 75.46 $\pm$ 0.58 & 68.86 $\pm$ 0.56 & 71.96 $\pm$ 4.29 & 67.88 $\pm$ 1.39 & 69.77 $\pm$ 1.56 \\
BioSentVec (PCA) & 65.37 $\pm$ 1.04 & 61.41 $\pm$ 0.82 & 68.45 $\pm$ 6.19 & 60.44 $\pm$ 1.82 & 64.01 $\pm$ 2.15 \\
\textbf{WellFactor\textsubscript{NS}} & \underline{81.47} $\pm$ 0.56 & \underline{73.68} $\pm$ 0.70 & \underline{74.70} $\pm$ 4.27 & \textbf{73.64} $\pm$ 2.98 & \underline{74.01} $\pm$ 0.81 \\
\textbf{WellFactor\textsubscript{S}} & \textbf{81.65} $\pm$ 0.55 & \textbf{73.98} $\pm$ 0.55 & \textbf{76.13} $\pm$ 1.91 & \underline{73.11} $\pm$ 0.97 & \textbf{74.57} $\pm$ 0.79 \\
\bottomrule
\end{tabular}
\end{table*}
\section{Evaluation and Results}
In this section, we evaluate our method across various tasks related to patient profiling and recommendation.
Specifically, our evaluation concentrates on three primary tasks: classification (identifying which group a patient belongs to), clustering (grouping similar patients), predicting patient embeddings, and similarity search (finding patients analogous to a given example).

\subsection{Evaluation Setting}
\subsubsection{Competing Methods}
We selected several baseline algorithms for user embedding as comparisons for our proposed method:
\begin{itemize}
\item \textbf{HashGNN~\cite{tan2020learning}:} Given that our problem setting can be conceptualized as graphs, incorporating HashGNN is a logical step. It involves constructing a graph where nodes signify patients, diagnoses, queries, and pages. This method was run with standard parameters.
\item \textbf{Text Embeddings:} \textit{GPT-2} and \textit{SentenceBERT} are prominent text embeddings. To create user embeddings with them, we computed embeddings for each domain-specific text associated with users and then derived their average.
\item \textbf{Biomedical-Domain Embedding:} \textit{BioSentVec}, specifically designed for the biomedical domain, was applied only to diagnosis domain texts to compute average embeddings.
\end{itemize}

These competitors were selected based on their common use and relevance to our application. 
Since GPT-2, SentenceBert, and BioSentVec were utilized to constitute our initial data features, our intent is to assess if an integrative embedding approach with the same embedding size can improve upon performance of these individual methods.

\subsubsection{Experimental Details}
For a fair comparison, we standardized the embedding length at 128 dimensions. If the inherent dimensionality of any method exceeded this, we employed Principal Component Analysis (PCA) to reduce the dimension to 128.
In our XGBoost model, we appended patients' age and gender information to the embedding for training and testing, acknowledging age and gender's potential impact on health-related outcomes. We treated gender as a categorical feature, exercising the flexibility XGBoost offers in handling diverse data types.
To ensure the reliability and stability of our results, we averaged metrics over 10 experiments and also computed the standard deviations. This iterative approach provides a broad overview of the method's robustness.

\begin{table*}[ht]
\centering
\caption{Summary of Cluster Analysis Results}
\label{tab:factor_analysis}
\begin{tabular}{r|c|c|c|c|c}
\toprule
\textbf{Data Domain} & \textbf{Cluster 1} & \textbf{Cluster 2} & \textbf{Cluster 3} & \textbf{Cluster 4} & \textbf{Cluster 5} \\
\midrule
\textbf{Search 1} & cpap & autopay & general & autopay & pharmacy \\
\textbf{2} & apria & auto & optometrist & auto & refill \\
\textbf{3} & supplies & taxes & mandarin & payments & refills \\
\textbf{4} & department & 1095a & indian & payonline & receipts \\
\textbf{5} & wang & payonline & appoitment & send & phone \\
\midrule
\textbf{Browse 1} & sleep & premium & task & html5 & technical \\
\textbf{2} & problems & estimates & whoops & reprom & dispose \\
\textbf{3} & durable & billing & ns & pb & unwanted \\
\textbf{4} & equipment & getting & videovist & vb & termsconditions \\
\textbf{5} & apnea & manager & footer & autopay & drugs \\
\midrule
\textbf{Diagnose 1} & sleep & bloating & nose & resuscitate & meningioma \\
\textbf{2} & apnea & salivary & dysphagia & myasthenia & myopathy \\
\textbf{3} & snoring & leiden & idiopathic & gravis & manic \\
\textbf{4} & cpap & splints & prolactinoma & macrocytosis & fall \\
\textbf{5} & rhythm & shin & hands & dependent & l1 \\
\bottomrule
\end{tabular}
\end{table*}
\subsection{Classification for App Recommendations}
In applying our method, our goal is to predict if a user will engage with a mental health support app among the available options. This task bears resemblance to link prediction or ranking challenges commonly encountered in the realm of information retrieval.
To assess the quality of our app recommendations, we collected visitation logs for the relevant download pages of the apps: Calm, Ginger, and myStrength as described in \autoref{subsec:label_collection}. A binary representation was employed to capture user engagement. Users visiting the download page at least once were assigned a value of 1, and all others were assigned a value of 0.
We employed XGBoost for our evaluations, due to its reputation for efficiency and accuracy. XGBoost, an implementation of gradient boosted decision trees, is known for its flexibility in handling various data types, including categorical variables like gender. Consequently, we incorporated patients' age and gender information in concatenation with the computed embeddings, widening data's breadth and depth.

\subsubsection{Performance Metrics}
To evaluate the classification performance, we compare the output predictions with actual labels.
We compute the count of categories: true positive (TP), false positive (FP), true negative (TN), and false negative (FN). Then, fundamental metrics such as recall (or true positive rate, TPR), precision, and the F1 score are computed.
The classifier's outcome is a list of probabilities of patients being positive. Setting the optimal threshold of the probability will modify the number of predicted positives and metrics like precision and recall.
The ROC curve visualizes this by plotting the TPR against the false positive rate (FPR) across thresholds between 0 and 1.
The area under this curve, known as the ROC-AUC, gives us a singular metric indicative of our model's performance across all thresholds.

To address the problem that metrics such as recall, precision or F1 score are sensitive to threshold selections, the Youden index offers a way to determine an optimal threshold. This index, mathematically represented as $J=TPR - FPR$, computes the threshold where the balance between sensitivity (or recall) and specificity is maximized. Thus, by optimizing the Youden index, we identify a threshold that best translates our continuous prediction scores into binary outcomes, enabling meaningful comparisons using metrics such as Precision, Recall, and F1 score.

\subsection{Classification results}
Our proposed methods, namely WellFactor\textsubscript{NS} (without semi-supervision) and WellFactor\textsubscript{S} (with semi-supervision), have showcased remarkable efficacy when contrasted with other baseline techniques. Specifically, the WellFactor\textsubscript{S} method achieved an impressive ROC-AUC of 81.65\%, accuracy of 73.98\%, recall of 76.13\%, and an F1-score of 74.57\%. Its counterpart, WellFactor\textsubscript{NS}, although slightly trailing in some metrics, presented a praiseworthy ROC-AUC of 81.47\% and an F1-score of 74.01\%, reflecting its competitive performance.

\subsection{Cluster Analysis of Patients}\label{subsec:cluster-analysis}

Our approach inherently provides information on both embedding and soft clustering results. One of the unique attributes of \method is the imposition of nonnegative constraints ~\cite{drake2010Raman}.
The dimension with the highest value in a patient's embedding can be interpreted as the predominant cluster that best describes that patient.

The basis matrices, represented by $\mathbf{W}_i (i=s,b,d)$, serve as the outputs of our model, characterizing representative patients for each cluster within specific domains (search, browse, and diagnose). 
It is important to note that while part of each $\mathbf{W}_i$ matrix is derived from term-frequency matrix decomposition, other rows correspond to other text embeddings. Every row in $\mathbf{W}_i$ represents either a specific keyword from the term-frequency decomposition or an embedding from the other mentioned models.

Using \method, we derived a 128-dimensional embedding. Out of these 128 clusters, we selectively emphasized the five largest clusters. Then, the top five keywords are selected within each cluster that exhibited the highest values in the corresponding columns of $\mathbf{W}_i$ matrices for each domain. The results are summarized in~\autoref{tab:factor_analysis}.
For example, in Cluster 1 in particular, we observe recurring keywords such as `sleep' and `apnea' across various domains, demonstrating the robustness of our method. Such findings underline the efficacy of our approach and also suggest its potential application in facilitating group-based recommendations.

\begin{table*}[ht]
\centering
\caption{
Performance of the proposed embedding method for predicting major diseases in Q4 based on data from Q1, Q2, and Q3. The evaluation is done by finding the top similar patients using the vector distance in the embedding space and checking how many of them are in the same cohort. The precision@k values indicate the proportion of true positive predictions among the top k predictions.
}
\label{tab:prediction-results}
\begin{tabular}{lrrrr|r}
\toprule
\textbf{Disease} & \textbf{Number of Patients in Cohort} & \textbf{precision@10} & \textbf{precision@20} & \textbf{precision@50} & \textbf{Random Precision} \\
\midrule
Hypertension & 3,244 & 1.25\% & 1.23\% & 1.18\% & 0.51\% \\
Diabetes & 5,313 & 2.78\% & 2.65\% & 2.49\% & 0.83\% \\
COPD & 363 & 0.34\% & 0.25\% & 0.20\% & 0.06\% \\
Chronic Kidney Disease & 2,949 & 2.67\% & 2.64\% & 2.41\% & 0.46\% \\
Depression & 2,945 & 1.28\% & 1.29\% & 1.27\% & 0.46\% \\
Cancers & 2,136 & 0.95\% & 0.91\% & 0.91\% & 0.34\% \\
Obesity & 6,159 & 2.35\% & 2.32\% & 2.31\% & 0.97\% \\
Osteoarthritis & 2,269 & 0.78\% & 0.80\% & 0.83\% & 0.36\% \\
\bottomrule
\end{tabular}
\vspace{-3mm}
\end{table*}

\subsection{Prediction and Similarity Search}
The subsequent analyses focus on the evaluation of our embedding method for predictability and similarity search. Given the multifaceted nature of healthcare data, it is critical to assess whether our embedding technique can capture the intrinsic relationships between patients and reliably predict their future interactions or diagnoses. Our primary objective is to evaluate the predictability of our embedding technique. Here is how we approached this:

\begin{table}[t]
\centering
\caption{Selected Diseases and their ICD-10 Codes}
\label{tab:disease_icd10}
\begin{tabular}{ll}
\toprule
Disease & ICD-10 Code \\
\midrule
Hypertension & I10 \\
Diabetes & E11, E12, E13, E14 \\
Chronic Obstructive Pulmonary Disease (COPD) & J44 \\
Chronic Kidney Disease & N18 \\
Depression & F32, F33 \\
Cancers & C00-C97 \\
Obesity & E66 \\
Osteoarthritis & M15-M19 \\
\bottomrule
\end{tabular}
\end{table}

\paragraph{Selection of Diseases for Analysis}
As presented in Table \ref{tab:disease_icd10}, the diseases chosen for this study were selected based on worldwide prevalence, profound impact on mortality, and significant associations with modifiable risk factors.
Hypertension and Diabetes, for instance, are directly influenced by dietary habits, sedentary lifestyles, and obesity~\cite{world2014global}.
COPD (Chronic Obstructive Lung Disease) emerges primarily due to smoking and environmental pollutants~\cite{singh2019global}, while Chronic Kidney Disease is frequently a complication of other conditions, illustrating the intricate interrelations of these diseases~\cite{levin2017global}. Depression, exacerbated by modern-day stressors and societal pressures, is becoming increasingly prevalent~\cite{world2017depression}. Cancers, on the other hand, span a vast range of conditions, with a mix of genetic, environmental, and lifestyle origins~\cite{sung2021global}.
Additionally, the study addresses obesity, with recent findings indicating alarming health effects of being overweight and obesity in a majority of countries over the past 25 years~\cite{gbd2017health}.
Osteoarthritis, a degenerative joint disease particularly of the hip and knee, represents another significant global health burden, and its occurrence is expected to increase with an aging population~\cite{cross2014global}.

\paragraph{Data Preparation}
We initially established cohorts of individuals diagnosed with major diseases in the 4th quarter (Q4) of 2022. The underlying assumption is that people with similar embedding vectors are more likely to share certain health outcomes, including the emergence of specific diseases.

\paragraph{Embedding Calculation}
Using data from the first three quarters (Q1, Q2, and Q3), we computed the embedding for each patient. This ensures that our embedding results are derived solely from historical data, allowing us to make predictions about events in the upcoming quarter (Q4).

\paragraph{Similarity Search}
For each patient diagnosed with a major disease in Q4, we calculated the vector distance between their embedding (from Q1-Q3) and the embedding results of all other patients. This allowed us to rank other patients based on their similarity.

\paragraph{Performance Assessment}
To evaluate the efficacy of our embedding method in identifying similar patients, we measured its accuracy using the ``precision@$k$" metric. This metric indicates the proportion of retrieved relevant patients among the top $k$ predictions. Given a patient diagnosed with a disease in Q4, another patient is considered ``relevant" if they were also diagnosed with the same disease in Q4. Formally:
$$\text{Precision@k} = \frac{\text{\# of relevant items in the top k predictions}}{k}.$$
As a baseline, we also compute the theoretical precision when matching patients randomly by ranking patient similarities without any consideration for their embedding vectors. Thus, we aim to demonstrate the absolute baseline performance one might expect without any predictive modeling.
Comparing the precision values obtained from our embedding approach with the theoretical random baseline offers a clear picture of our method's predictive capability.

\paragraph{Discussion of Results}
The results, as summarized in ~\autoref{tab:prediction-results}, clearly demonstrate the potential of our embedding approach for predicting major diseases for patients based on historical data.
For every disease under consideration, our method consistently outperforms the theoretical random baseline, highlighting its efficacy in capturing meaningful patient similarities. Furthermore, even in instances where the precision values might seem modest, such as in COPD, the difference between our method's precision and the random baseline is still marked, reinforcing the utility of our approach for identifying patients with similar health outcomes. 
\section{Conclusion}
We propose \method, a method for patient profiling using integrative embedding of healthcare data. 
Our method seamlessly integrates information from multiple sources to create comprehensive patient profiles and has been shown to outperform existing methods in classification, clustering, and similarity searches. It efficiently handles missing or unobserved data and can compute embeddings for previously unseen patients. \method is a versatile tool for deriving meaningful insights from diverse healthcare data sources, enabling personalized healthcare recommendations and improved patient care.
\section*{Acknowledgment}
This work was funded by Kaiser Permanente, and we extend our gratitude to Ji Yeon Kim for her assistance with the illustrations.
\bibliographystyle{IEEEtran}
\bibliography{bibs}
\end{document}